\documentclass[journal]{IEEEtran}
\usepackage{ifpdf}
\usepackage{cite}
\ifCLASSINFOpdf
   \usepackage[pdftex]{graphicx}
   \graphicspath{{../pdf/}{../Images/}}
   \DeclareGraphicsExtensions{.pdf,.jpg,.png}
\else
\fi
\usepackage{float}

\usepackage{amsmath}
\usepackage{algorithmicx}
\usepackage{algpseudocode}

\usepackage{array}
\ifCLASSOPTIONcompsoc
  \usepackage[caption=false,font=normalsize,labelfont=sf,textfont=sf]{subfig}
\else
  \usepackage[caption=false,font=footnotesize]{subfig}
\fi
\usepackage{fixltx2e}
 \usepackage{dblfloatfix}
\usepackage{url}
\begin{document}
\title{An Efficient Approach for Object Detection and \\ Tracking of Objects in a Video with Variable Background}

\author{Kumar~S.~Ray
        and~Soma~Chakraborty
        % <-this % stops a space
\thanks{This work is a sponsored project of Indian Statistical Institute, Kolkata, India.}
\thanks{Kumar S. Ray is with Indian Statistical Institute, 203 B.T.Road, Kolkata-108, India 
					(e-mail: ksray@ isical.ac.in).}% <-this % stops a space
\thanks{Soma Chakraborty is with Indian Statistical Institute, 203 B.T.Road, Kolkata-108, India 
					(e-mail: soma.gchakraborty@gmail.com).}}

\maketitle

\begin{abstract}
This paper proposes a novel approach to create an automated visual surveillance system which is very efficient in detecting and tracking moving objects in a video captured by moving camera without any apriori information about the captured scene. Separating foreground from the background is challenging job in videos captured by moving camera as both foreground and background information change in every consecutive frames of the image sequence; thus a pseudo-motion is perceptive in background. In the proposed algorithm, the pseudo-motion in background is estimated and compensated using phase correlation of consecutive frames based on the principle of Fourier shift theorem. Then a method is proposed to model an acting background from recent history of commonality of the current frame and the foreground is detected by the differences between the background model and the current frame. Further exploiting the recent history of dissimilarities of the current frame, actual moving objects are detected in the foreground. Next, a two-stepped morphological operation is proposed to refine the object region for an optimum object size. Each object is attributed by its centroid, dimension and three highest peaks of its gray value histogram. Finally, each object is tracked using Kalman filter based on its attributes. The major advantage of this algorithm over most of the existing object detection and tracking algorithms is that, it does not require initialization of object position in the first frame or training on sample data to perform. Performance of the algorithm is tested on benchmark videos containing variable background and very satisfiable results is achieved. The performance of the algorithm is also comparable with some of the state-of-the-art algorithms for object detection and tracking.
\end{abstract}


% \fi
\IEEEpeerreviewmaketitle

\section{Introduction}
\IEEEPARstart{M}{ethods} of extracting significant information from still images and videos captured in a constrained environment, are being studied for several decades to enhance images or to build automated systems. But in the recent years, increased demand of intelligent automated systems necessitated the processing of more challenging real world scenes which possess many complexities; like noise in data, abrupt motion or illumination variation, non-rigid or articulated movement of objects, background variation etc. In most of the intelligent automated systems like surveillance systems, intelligent vehicles  and so on, real world image sequences are processed to detect and track  dynamic objects in complex environment as an initial step. The rise of advanced computational systems and efficient imaging devices have facilitated to capture noise-free images and process high-dimensional data very efficiently and many benchmark methods have been established for detecting and tracking  dynamic objects in complex environment \cite{milan2016mot16}, \cite{wu2015object}, \cite{kristan2015visual}. Still, object detection and tracking in videos with variable background along with other complexities pose a serious challenge and has emerged as a rigorous research problem. When a video is captured by a camera installed on a non-static surface (say, moving vehicle), each pixel of a pair of consecutive frames contains different information, giving an impression of motion in background pixel too. Thus it is challenging task to separate a pixel containing background information from a pixel containing foreground information; as usual methods of extracting foreground using a background template is infeasible for videos with variable background. Active researches are being continued and some works have been reported so far, on successful detection and tracking of moving objects in variable background \cite{chuang2016underwater}-\cite{tang2016multi}, \cite{meier2015detection}-\cite{markovic2014moving} and \cite{arvanitidou2013motion}-\cite{kratz2012tracking}. Most of these algorithms calculate optical flows or detect feature points or region of interest in frames and estimate global or camera motion by comparing point trajectories or optical flows with geometric or probabilistic motion model. Moving objects are detected by extracting classified foreground features or removing background features. 

However, these methods process spatial information to estimate pseudo motion or to classify feature points as background or foreground. In our present work, we have estimated and compensated pseudo motion by frequency domain analysis of video frames. In this approach, Fourier transformed form of a pair of consecutive frames are analyzed as a whole to estimate their relative shift due to camera movement and the translation is also compensated in frequency domain applying the principle of Fourier shift theorem. Thus, this approach is more efficient than spatial domain analysis as it reduces the computational complexity due to feature point detection and tracking or optical flow computation and matching to estimate the inter-frame translation offset. We have devised a method to separate background and foreground information and remove flickering background or noise, by analyzing  the pseudo motion compensated current frame and few of its preceding frames. We have also formulated a method of morphological operation to refine the region of detected moving objects. Finally we represented each detected object by its centroid, dimensions and intensity distribution and tracked it through Kalman filtering on its features. Contributions of this paper are:
\begin{itemize}
	\item The proposed algorithm applied the principle of Fourier shift theorem and phase correlation method very effectively to efficiently estimate and compensate pseudo motion in background due to camera movement.
	\item A method is devised to model an acting background by exploiting the history of commonality of a frame and detect foreground. Actual moving objects are detected in foreground by removing flickering background or noise using recent history of dissimilarity of the frame. Another method of morphological operation is also presented in the proposed algorithm to refine object regions.
	\item The proposed algorithm detects object in variable background without prior knowledge of environment or shape of objects or additional sensor information. It demonstrated satisfactory performance on benchmark dataset \cite{wu2015object}, \cite{kristan2015visual} containing  videos with variable background without manual initialization of object region or training on sample data. Performance of the proposed algorithm is also comparable with state-of-the-art methods \cite{duffner2016using}, \cite{henriques2015high}, \cite{maresca2014clustering}.
\end{itemize}

The rest of the paper is organized as follows: the related works on detection and tracking of moving objects in variable background are described in section II. The proposed method is elaborated in Section III. Algorithm and its computational complexity are described in section IV. Experimental results are presented in section V followed by conclusion and future work in section VI.

\section{Related Works}
Algorithms which detect and track moving objects in videos captured by moving camera, handled the challenge of separating background and foreground information; by analyzing one or more sensor data statistically or by fitting image information with some geometric or probabilistic motion models. 

In advanced driver assistance systems, moving object is detected by data fusion from various movable sensors like lidar (Light Detection And Ranging), camera, radar etc. As in \cite{chavez2016multiple}, lidar data is used to populate Bayesian occupancy grid map to locate vehicles by a Maximum Likelihood approach on the grid map. Visual descriptors of objects are generated using Sparse Histograms of Oriented Gradients descriptor (S-HOG) on camera images. Feeding these information to the radar sensor a target is tracked using Interactive Multiple Model (IMM). In \cite{hwang2016fast}, off-the-shelf algorithms are used for object detection in color image (2D) and lidar data (3D) space by extracting local and global histograms. Objects are classified using linear SVM and tracked by segment matching based method. However, use of additional sensors limits the ease of application of \cite{chavez2016multiple} and \cite{hwang2016fast}. In \cite{chuang2016underwater}, an object is configured in a set of deformable kernels and featured by histograms of color, texture and oriented gradients (HOGs). Kernel motion is computed by maximizing the color and texture histograms similarity between a candidate and target model using mean-shift algorithm. Configuration of object part is restored by optimizing the part deformation cost using mean-shift algorithm on HOG feature. In \cite{xiao2016track}, object appearance and motion proposal scores are calculated from color and optical flow magnitude map. Similar object proposals with high scores are clustered. Temporal consistency of each cluster of a frame is calculated by Support Vector Machine (SVM) detector. Cluster with highest detection score in each frame is added repetitively to generate a spatio-temporal tube for an object. In \cite{tang2016multi}, detection hypotheses for a moving object are generated using a state-of-the-art object detector. Detection hypotheses within and between frames are connected as graph. Affinity measures for pairs of detections in two consecutive frames are estimated by 'Deep Matching' through multi-layer deep convolutional architecture. The graph is partitioned into distinct components (object's track) by solving the minimum cost sub-graph multicut problem. 

In \cite{meier2015detection}, moving objects are detected as outliers of optical flow (OF) measurements by estimating the ego-motion using the linear and angular velocity of the aerial camera. In another system for advanced vehicle \cite{asvadi2015detection}, authors represented the dynamic environment of each frame through 2.5D map of sensor measurements- the point cloud updated with localization data and no ground cell (cell value with low variance and height). In every frame, moving objects are extracted from 2.5D map using spatial reasoning. Detected moving objects are tracked by applying data association and Kalman filtering. In \cite{hou2015deformable}, kernels of a Deformable Part Model (DPM) in a frame are mean-shifted based on spatially weighted color histogram to the new temporal locations. Part configuration is maintained by applying deformation costs statistically inferred from mean-shift on histogram of oriented gradient (HOG) features of the current frame. In \cite{hu2015moving} and \cite{hu2014effective}, feature points in the frames are extracted using off-the-shelf algorithms and are classified as foreground or background points by comparing them with multiple-view geometry. Foreground regions are obtained through image differencing and integrating foreground feature points. Moving objects are detected by applying the motion history and refinement schemes on the foreground. In \cite{hu2015moving}, moving object is tracked using a Kalman filter based on the center of gravity of a moving object region. In \cite{fragkiadaki2015learning}, promising object proposals are computed using multiple figure-ground segmentation on optical flow boundaries and are ranked with a Moving Objectness Detector (MOD) trained on image and motion fields. Top ranked segments in each frame are extended into spatio-temporal tubes using random walkers on motion affinities of dense point trajectories. 

In \cite{ferone2014neural}, adaptive neural self-organizing background model is generated to automatically adjust the background variations in each frame of video sequences captured by a pan-tilt-zoom (PTZ) camera. However, the center of a PTZ camera is fixed and gives maximum 360 degree view of a particular region; thus providing favorable condition for creating a background model. But a camera mounted on a mobile platform captures a wide range of scene for which modeling a background is much more challenging. Zamalieva et al. estimated geometric transformations between two consecutive frames through dense motion fields using Geometric Robust Information Criterion (GRIC) \cite{zamalieva2014multi}. Appearance models are propagated from previous to current frame using selected geometric transformation and the fundamental matrix estimated by a series of homography transforms. Background/ foreground labels are obtained by combining motion, appearance, spatial and temporal cues in a maximum-a-posteri Markov Random Fields (MAP-MRF) optimization framework. In \cite{zhang2014fast}, the target location is initialized manually or by some object detection algorithms in the first frame. Spatial correlations between the target and its neighborhood is learned and is used to update a spatio-temporal context model for the next frame. The object location likelihood or confidence map in a frame is formulated using the prior information of the target location, scale parameter and shape parameter. Object is tracked by finding maximum value in the confidence map. In a robotic system \cite{markovic2014moving}, a frame is represented on the unit sphere to categorize sparse optical flow (OF) features as dynamic or static points by analyzing the distance of terminal vector points to the great circle arc. The dynamic flow vectors constitute an object and is tracked using a directional statistics distribution on the unit sphere. 

The authors of \cite{arvanitidou2013motion}, compensated global motion in frames using the Helmholtz Tradeoff Estimator and two motion models. Error maps are generated by fusing compensated frames bidirectionally and applying the compensated motion vector fields. Connectivity of high error values are established by hysteresis thresholding with optimal weight selected using weighted mean. In \cite{choi2013general} observation cues are generated using a three-dimensional (3D) coordinate system. Tracking is solved by finding maximum a posteri (MAP) solution of a posterior probability and the reversible jump Markov chain Monte Carlo (RJ-MCMC) particle filtering method. However, use of a depth sensor to generate 3D cues limits its applications. In \cite{lian2013voting}, ego-motion is estimated and compensated by applying voting decision on a set of motion vectors, determined by the edge feature of objects and/or background. Ego-motion compensated moving edges are corrected and enhanced by morphological operations to construct moving objects. Zhou et al. compensated camera motion using a parametric motion model and non-convex penalty and applied Markov Random Fields (MRFs) to detect moving objects as outliers in the low-rank representation of vectorized video frames in \cite{zhou2013moving}. However, performance of this method in convergence to a local optimum depends on initialization of foreground support. Also it is not suitable for real-time object detection as it works in a batch mode. In \cite{kim2013detection}, moving objects are separated from background based on pixel-wise spatio-temporal distribution of Gaussian on non-panoramic adaptive background model. Camera motion is estimated by applying Lucas Kanade Tracker (LKT) to edges in current frame and background model. Moving objects have been tracked in H.264/AVC-compressed video sequences using an adaptive motion vectors and spatio-temporal Markov random field (STMRF) model in \cite{khatoonabadi2013video}. Target object is selected manually in the first frame and in each subsequent frames object motion vectors (MVs) are calculated through intracoded block motion approximation and global motion (GM) compensation. Rough position of the object in the current frame is initialized by projecting its previous position using the estimated GM parameters and then MVs are updated by STMRF. 

In \cite{lim2012modeling}, temporal model propagation and spatial model composition are combined to generate foreground and background models and likelihood maps are computed based on the models. Kratz et al. proposed an idea to model crowd motion using a collection of hidden Markov models trained on local spatio-temporal motion patterns in \cite{kratz2012tracking}. Prediction from this motion model is feed to a particle filter to track spatio-temporal pattern of an individual in videos.

\section{Proposed Method}
\newcommand\inv[1]{#1\raisebox{1.15ex}{$\scriptscriptstyle-\!1$}}
\subsection{Estimation and compensation of pseudo motion in  background}
\begin{figure}[t]
\centering
\subfloat[]{\includegraphics[width=1in]{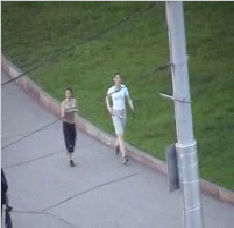}
\label{fig_1a}}
\subfloat[]{\includegraphics[width=1in]{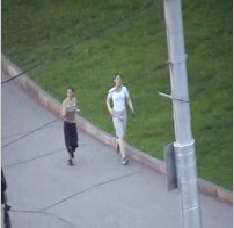}
\label{fig_1b}}
\subfloat[]{\includegraphics[width=1in]{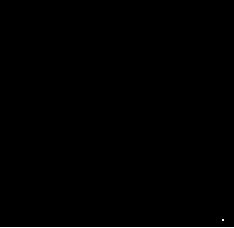}
\label{fig_1c}}
\caption{a. A Frame in an image sequence. b. A translated version of the frame. c. Spatial representation of phase correlation- second image is translated by the white pixel at the bottom-right corner.}
\label{fig_1}
\end{figure}

 In a video captured by a moving camera both background and foreground information change in each consecutive frame due to camera movement i.e. same pixel in two consecutive frames contains different intensity values. But, change in background pixel values are different in nature with respect to the change in foreground pixels values. Background pixels of a frame are translated in the direction of the camera by the same amount as per the displacement of camera. But, the foreground pixels are translated in the direction of object's movement by the amount of relative displacement of object and the camera. Thus, if relative global displacement due to camera movement between two consecutive frames of a video are estimated; we can compensate the pseudo motion in background pixels of a frame. In our work, we have employed phase correlation to calculate the relative translative offset to estimate global displacement between consecutive video frames, as frequency-domain analysis is more resilient than spatial-domain approach in presence of noise, occlusion etc. It is apparent that, in a video captured by a moving camera each successive frame is a linearly shifted version of the previous one. That is if $v_t(x,y)$ is a video frame at $t^{th}$ time instance and $v_{t+1}(x,y)$ is it's successive frame at time instance $(t+1)^{th}$, then $v_{t+1}(x,y)=v_t(x-\Delta x, y-\Delta y)$.

As per Fourier-shift theorem, spatial shift in an image with respect to another similar image results into a global phase shift in frequency domain. That is if $V_t(m,n)=\mathcal{F}\{v_t(x,y)\}$ and $V_{t+1}(m,n)=\mathcal{F}\{v_{t+1}(x,y)\}$ are frequency domain representatives of frames at $t^{th}$ and $(t+1)^{th}$ time instances then, 
\begin{equation}
\label{eqn_2}
V_{t+1}=V_t e^{-i2\pi(\frac{m\Delta x}{M}+\frac{n\Delta y}{N})}
\end{equation}
where, $(M,N)=$ dimension of a frame and $(\Delta x,\Delta y)=$ relative spatial displacement between two consecutive frames which is to be estimated.

The normalized cross-power spectrum calculated by multiplying of $V_t(m,n)$ and complex conjugate of $V_{t+1}(m,n)$ factors out the global phase difference between two successive frames:
\begin{align*}
P(m,n)&=\frac{V_t V_{t+1}^*}{|V_t V_{t+1}^*|} \\
    &=\frac{V_t V_t^* e^{-i2\pi(\frac{m\Delta x}{M}+\frac{n\Delta y}{N})}}{|V_t V_t^* e^{-i2\pi(\frac{m\Delta x}{M}+\frac{n\Delta y}{N})}|} \\
    &=\frac{V_t V_t^* e^{-i2\pi(\frac{m\Delta x}{M}+\frac{n\Delta y}{N})}}{|V_t V_t^*|} 
\end{align*} 
\begin{equation} 
\label{eqn_3}
P(m,n)=e^{-i2\pi(\frac{m\Delta x}{M}+\frac{n\Delta y}{N})}
\end{equation}

Inverse Fourier transform of the complex exponential (P) is an impulse of single peak in spatial domain as shown in fig. \ref{fig_1c}: 
\begin{equation} 
\label{eqn_4}
\mathcal{\inv F}\{P(m,n)\}=\delta(x-\Delta x, y-\Delta y)
\end{equation} 
where, the location $(x-\Delta x, y-\Delta y)$ is the position of maximum global displacement in frame $v_{t+1}$ with respect to frame $v_t$. We have calculated the amount of spatial shift in horizontal and vertical direction as per eq. (\ref{eqn_5}) and (\ref{eqn_6}):
\begin{equation}
\label{eqn_5}
\Delta x = \begin{cases}
      \overline{x}-M-1, & \text{for $\overline{x}>M/2$} \\
      \overline{x}-1, & \text{otherwise}
    \end{cases}
\end{equation}
\begin{equation}
\label{eqn_6}
\Delta y = \begin{cases}
      \overline{y}-N-1, & \text{for $\overline{y}>N/2$} \\
      \overline{y}-1, & \text{otherwise}
    \end{cases}
\end{equation}
where, $\overline{x}=x-\Delta x$ and $\overline{y}=y-\Delta y$. This amount of spatial shift is used to compensate the global displacement of all the pixels in the frame $v_{t+1}$ with respect to $v_t$ by applying the Fourier shift theorem on $V_{t+1}$:
\begin{equation} 
\label{eqn_7}
\overline{v_{t+1}}=\mathcal{\inv F}\{V_{t+1} e^{-i2\pi(\frac{m\Delta x}{M}+\frac{n\Delta y}{N})}\}
\end{equation}

In our work, we have formulated a method to construct an acting background model for the current frame from recent history of the frame. So, in this part of the work we have estimated and compensated the pseudo motion of the pixels in $\eta$ number of previous frames of the current frame ($v_t$), where $\eta << T$  = length of the image sequence. That is, frames $v_{t-1}...v_{t-\eta}$ are modified to compensate the global displacement with respect to $v_t$ and resulted in $\overline v_{t-1}...\overline v_{t-\eta}$ as recent history of the current frame $v_t$.

\subsection{Modeling an acting background for $t^{th}$ frame}
\begin{figure}[t]
\centering
\includegraphics[width=3in]{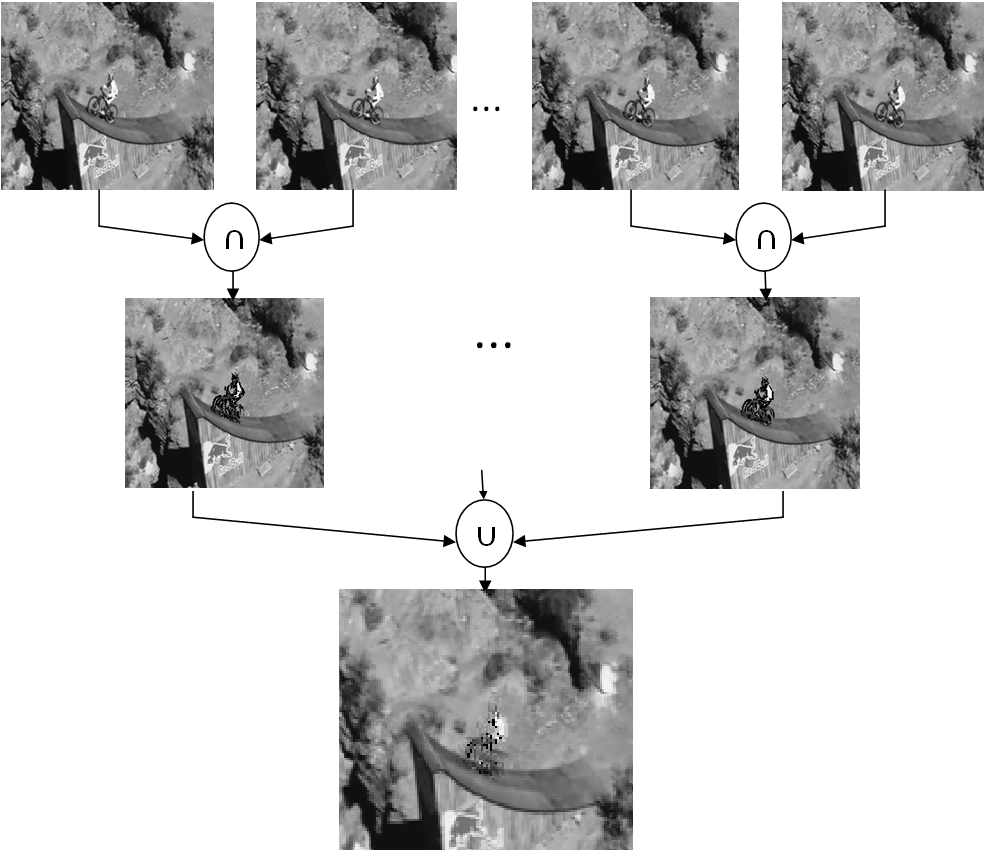}
\caption{Formation of an acting background for a frame in an image sequence}
\label{fig_2}
\end{figure}

The motivation to model an acting background is the unrestricted movement of an articulated object like human. It is observed that, a non-rigid object may has a full body movement in each consecutive frames of a sequence or it may has only partial body movement like head or hand movement in few subsequent frames. So, searching local movement between only two consecutive frames may not result in precise shape and size of the moving object. On the other hand, finding difference of local information between few preceding frames and the $t^{th}$ frame has high probability to produce the whole region of a moving object. Considering immediate history of the $t^{th}$ frame also has the advantage in reducing the affect of abrupt illumination variation, partial occlusion and sudden change of object's velocity and direction. So, we have generated the acting background of the current frame by unifying the commonality of information in its history frames as described in the fig. \ref{fig_2}. As described in the previous section, recent history of the current frame ($v_t$) comprises of $\eta$ number of preceding frames- $\overline v_{t-1}...\overline v_{t-\eta}$. For the ease of expression; let, $H_1=\overline v_{t-1}, H_2=\overline v_{t-2},...,H_\eta=\overline v_{t-\eta}$. Each of the history frames ($H_i$) is quantized to map all the intensity values of range $0$ to ($2^8-1$) to a smaller set of values- $1$ to $10$: 
\begin{equation} 
\label{eqn_8}
H^{Q}_{i}=\{h^{Q}_{l}:h^{Q}_{l}=q_j*10, \text{for $q_{j-1}<h^{N}_{l}\leq q_j$}\}       
\end{equation}
where, $h^{N}_{l}\in \frac{H_i}{(2^8-1)}$ is a normalized value of a pixel at location $l$ in $H_i$ and $q_j\in [0.1,0.2,0.3,...,1]$ is a pivot on the quantization scale.

Then each pair of successive history frames ($H_i$)'s are intersected; to accumulate the common intensity values of background pixels in both of them by comparing their respective quantized forms ($H^{Q}_{i}$)'s: 
 \begin{equation} 
\label{eqn_9}
H_i \cap H_{i+1}=\{h_l:h_l=\begin{cases}
      \frac{h^{i}_{l}+h^{i+1}_{l}}{2}, & \text{for $|\overline{h}^{i}_{l}-\overline{h}^{i+1}_{l}|\leq 1$} \\
      0, & \text{otherwise}
    \end{cases}\}       
\end{equation}
where, $h^{i}_{l}\in H_i$, $h^{i+1}_{l}\in H_{i+1}$, $\overline{h}^{i}_{l}\in H^{Q}_{i}$ and $\overline{h}^{i+1}_{l}\in H^{Q}_{i+1}$ are intensities at the same location $l$ in the respective images. The pixels which are part of consistent background, produce common information in all the intersected images. On the other hand, pixels which are part of background, but are covered by a moving object are considered as part of foreground in prior history frames and eventually are absent in prior intersected images. As these pixels are gradually revealed in subsequent frames, they contribute in common information of posterior intersected images. The same phenomena happens when the background pixels are gradually being covered in most recent history frames. To collect common information from recent history frames as much as possible, intersected images are unified to generate the acting background model for the current frame ($v_t$):
 \begin{equation} 
\label{eqn_10}
B_t=\bigcup\limits_{i=1}^{\eta-1}(H_i \cap H_{i+1})
\end{equation} 
We have also analyzed the history of dissimilarities of prospective foreground pixels in $v_t$ as follows: 
\begin{equation} 
\label{eqn_11}
D=\frac{1}{\eta}\sum^{\eta}_{i=1}D^{'}_{i}       
\end{equation}
where, $D^{'}_{i}$ is the absolute difference between prospective foreground pixels of each pair of history frames:
\begin{equation} 
\label{eqn_12}
D^{'}_{i}=\{d_l : d_l=
\begin{cases}
 |h^{i}_{l}-h^{i+1}_{l}|, & \text{for $|\overline{h}^{i}_{l}-\overline{h}^{i+1}_{l}|>1$} \\
      0, & \text{otherwise}
    \end{cases} \}            
\end{equation}
where, $h^{i}_{l}\in H_i$, $h^{i+1}_{l}\in H_{i+1}$, $\overline{h}^{i}_{l}\in H^{Q}_{i}$ and $\overline{h}^{i+1}_{l}\in H^{Q}_{i+1}$ are intensities at the same location $l$ in the respective images. 

We have given a weight to each pixel of $v_t$ based on its frequency of commonality in the recent history:
\begin{equation} 
\label{eqn_13}
W=\{w_l : w_l=
\begin{cases}
 w_l+1, & \text{for $|\overline{h}^{i}_{l}-\overline{h}^{i+1}_{l}|>1$} \\
      w_l, & \text{otherwise}
    \end{cases} \}            
\end{equation}
where, $\overline{h}^{i}_{l}\in H^{Q}_{i}$ and $\overline{h}^{i+1}_{l}\in H^{Q}_{i+1}$ are intensities at the same location $l$ in the respective quantized version of history frames. Higher the weight of a pixel, greater its probability of being a background pixels. It is observed in test image sequences that, a pixel having the highest weight is obviously a background pixel. On the other hand, a pixel with lowest weight ($w_l\rightarrow 0$) is an unstable pixel which usually belongs to flickering background like water, tree leaves etc. The pixels with weight in-between are either in gradually revealed or covered part of background or inside a large object area which have comparatively lesser displacement than the object edges.

\subsection{Detecting moving object in foreground}
\begin{figure}[t]
\centering
\includegraphics[width=2.5in]{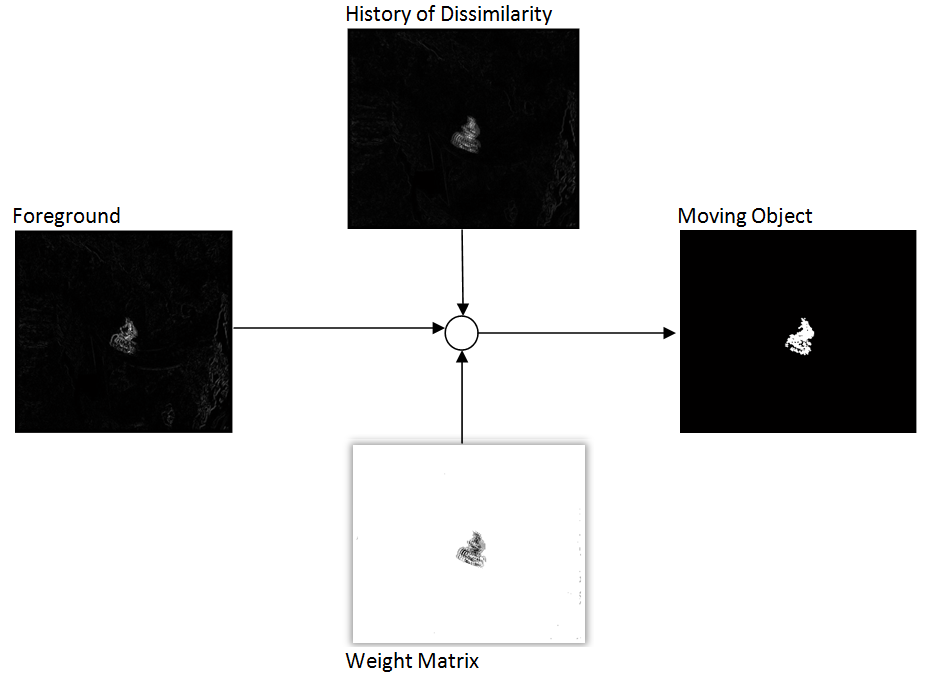}
\caption{Detecting moving object in foreground using history of dissimilarity and weight matrix}
\label{fig_3}
\end{figure}

Both weight matrix ($W$) and history of dissimilarity ($D$) of $v_t$ are divided in three levels- high, medium and low. For all weights in $W$- $W_{high}=\{w: w=(\eta-1)\}$ and $W_{low}=\{w:w\leq \lfloor\frac{(\eta-1)}{3}\rfloor\}$. Thus, $W_{medium}=\{w:\lfloor\frac{(\eta-1)}{3}\rfloor<w<(\eta-1)\}$, where $\eta$ is the number of history frames of $v_t$. Similarly, for each difference ($d\in D$), we define the three levels as: $D_{low}=\{d:d\leq\lfloor\frac{\chi}{3}\rfloor\}$, $D_{high}=\{d:d\geq\lfloor\frac{2\chi}{3}\rfloor\}$ and $D_{medium}=\{d:\lfloor\frac{\chi}{3}\rfloor<d<\lfloor\frac{\chi}{3}\rfloor\}$, where $\chi=(2^8-1)$ is the highest gray value. Pixels with weight in $W_{high}$ re assigned the zero value in both the acting background ($B_t$) and the current frame ($v_t$). Then the foreground is detected by simple frame differencing between the modified $B_t$ and the current frame $v_t$:
\begin{equation} 
\label{eqn_14}
F_t=|v_t-B_t|
\end{equation}

The foreground ($F_t$) contains the region of motion as the difference of intensities between background and the current frame. $F_t$ is analyzed further to mark each pixel as part of an actual moving object or a part of flickering background. Weight of each foreground pixel ($\rho$) plays an important role to decide the nature of the $\rho$. Motion history ($d\in D$) of a $\rho$ is also used to measure the strength f motion energy of the $\rho$. For the ease of comparison with its history of motion energy, all the pixels of foreground are also categorized as low, medium and high- $F_{low}=\{f:f\leq\lfloor\frac{\chi}{3}\rfloor\}$, $F_{high}=\{f:f\geq\lfloor\frac{2\chi}{3}\rfloor\}$ and $F_{medium}=\{f:\lfloor\frac{\chi}{3}\rfloor<f<\lfloor\frac{2\chi}{3}\rfloor\}$, where $\chi=(2^8-1)$ is the highest gray value. If weight of a pixel is medium and if its motion energy and history of motion are comparable; then that $\rho$ is a candidate part of an actual moving object. On the other hand, if a pixel has low weight then considering the level of its motion energy history the $\rho$ is marked as part of flickering background or part of an actual moving object:
\begin{equation} 
\label{eqn_15}
f_\rho=
\begin{cases}
 1, & \text{for $w_\rho\in W_{medium}$ \& $L(f_\rho)\geq L(d_\rho)$} \\
 0, & \text{for $w_\rho\in W_{low}$ \& $(d_\rho\in D_{high} \parallel d_\rho\in D_{medium})$} \\
 1, & \text{for $w_\rho\in W_{low}$ \& $d_\rho\in D_{low}$ \& $L(f_\rho)> L(d_\rho)$}
    \end{cases} \}            
\end{equation} 
where $L$ is the level of $\rho$. Combining all the $f_\rho$'s calculated in (\ref{eqn_15}), a modified foreground is generated which contains only the blobs of prospective moving objects in a variable background as shown in fig. \ref{fig_3}. 

\begin{figure}[t]
\centering
\subfloat[]{\includegraphics[width=0.8in]{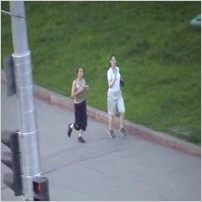}
\label{fig_4a}}
\subfloat[]{\includegraphics[width=0.8in]{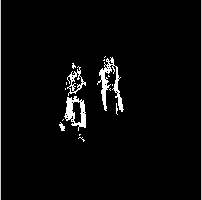}
\label{fig_4b}}
\subfloat[]{\includegraphics[width=0.8in]{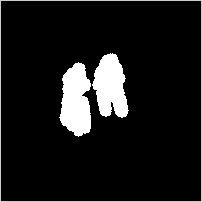}
\label{fig_4c}}
\subfloat[]{\includegraphics[width=0.8in]{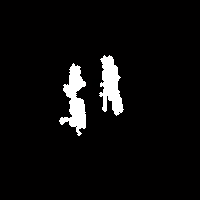}
\label{fig_4d}}
\caption{a. A Frame in an image sequence. b. Detected moving object. c. Dilated form of the detected moving object. d. Refined form of the moving object.}
\label{fig_4}
\end{figure}

Each of the blobs of prospective moving objects is further processed morphologically to refine the area of the moving objects. First, boundary of each blob (object) is dilated:
\begin{equation} 
\label{eqn_16}
O^{'}=O\cup\{\rho: \rho \in \alpha R \text{ and } (\sqrt{\rho^2-o^{2}_{hr}} \parallel \sqrt{\rho^2-o^{2}_{vr}})\leq d/2\}
\end{equation} 
where, $R=$ the region containing object $O$ and $\alpha$ is a constant set to $1.5$ from experience, $o_{hr}\in O$ and $o_{vr}\in O$ are nearest boundary pixel of $\rho$ on horizontal straight line and  vertical straight line respectively and $d=$ smallest distance between centroid and the boundary pixels of $O$. The dilated form of an object in fig. \ref{fig_4b} is depicted in fig. \ref{fig_4c}. The dilation may lead to merging of very closely spaced objects. But, this has the advantage of connecting the fragmented parts of an object and closing of holes inside the object. We have searched for pixels with significant local intensity changes in the dilated blobs/objects as expressed in (\ref{eqn_17}). These pixels usually falls on the edge of an object.
\begin{equation} 
\label{eqn_17}
E=\{\rho: \rho=(r,c)\in O^{'} \text{ and } max(I_{\overline \rho})-min(I_{\overline \rho})\geq \sigma(O)\}
\end{equation} 
where, $I_{\overline \rho}$ is the intensity at a location $\overline \rho \in [(r\pm 1,c),(r, c\pm 1),(r\pm 1,c\pm 1),(r\pm 1, c\mp 1)]$ and $\sigma(O)=$ standard deviation of all intensities of $O$. Then each dilated blob is scanned horizontally, to find any white pixel falling beyond the edge pixel in a row of $O^{'}$ and is removed from the blob:
\begin{equation} 
\label{eqn_18}
O_h=O^{'} -\{\rho_{hr}: \rho_{hr}<e^{first}_{hr} \parallel \rho_{hr}>e^{last}_{hr}\}
\end{equation} 
where, $hr$ denotes same row in $O^{'}$ and $E$ and $e_{hr}\in E$. Similarly, any white pixel beyond edge pixel in vertical direction is also removed from $O^{'}$:
\begin{equation} 
\label{eqn_19}
O_v=O^{'} -\{\rho_{vr}: \rho_{vr}<e^{first}_{vr} \parallel \rho_{vr}>e^{last}_{vr}\}
\end{equation} 
where, $vr$ is the same column in $O^{'}$ and $E$ and $e_{vr}\in E$. Finally, the refined object region is generated by:
\begin{equation} 
\label{eqn_20}
\overline O=O_h \cap O_v
\end{equation}  
The refined object region of an object is shown in fig. \ref{fig_4d}. Centroid and dimensions of an moving object is calculated to use while tracking the object as described in the next section. The centroid of the object detected in the current frame is searched in a space of following frame defined as: $S=\alpha (\widehat{h}\ast \widehat{w})$, where $\widehat{h}$ and $\widehat{w}$ are the height and width an object and $\alpha$ is a constant set as $1.5$ from experience. We have also used the gray value histogram of the moving object to solve any conflict in data association during tracking. But, the number of gray intensities to represent the color distribution of an object precisely, is a matter of deep consideration. Intuitively we can say that, within a region containing a single object, color variation is not high. So, if we calculate gray value histogram of all possible intensities for an object, a very small set of values is sufficient to describe the intensity distribution of the whole object. By experiments on the test videos, we have observed the validity of this intuition. So, in the present work we have calculated gray value histogram of all $(2^8-1)$ intensities. Then selected three highest peaks i.e. first to third highest peak of the histogram to represent the intensity distribution of an individual object. Thus, the feature vector of a moving object consists of: 1) centroid ($[\widehat{r},\widehat{c}]$), 2) dimensions ($[\widehat{h},\widehat{w}]$) and 3) intensity distribution ($[\widehat{g_1},\widehat{g_2},\widehat{g_3}]$) of the object.

\subsection{Tracking objects}
Tracking an moving object in variable background is to detect the object in the current frame and associate it with precisely the same object detected in the immediately successive frame. In our present work, we have employed Kalman filter \cite{kalman1960new} to generate the object trajectory. In each and every frame of an image sequence, each detected moving object is assigned a track consisted of the following fields: 1) object identifier (id), 2) dimensions of the object- $[\widehat{h},\widehat{w}]$, 3) intensity distribution- $[\widehat{g_1},\widehat{g_2},\widehat{g_3}]$, 4) Kalman filter, 5) the number of consecutive frames for which the object is not detected or the track is invisible- $\widehat {vc}$. The Kalman filter field of the track consists of  state vector parametrized as: $s=(\widehat{r},\widehat{c},\widehat {v_r},\widehat {v_c})$, where $[\widehat{r},\widehat{c}]=$ centroid of the object and $[\widehat {v_r},\widehat {v_c}]=$ horizontal and vertical components of velocity of the object. In our work we have assumed a constant velocity or zero acceleration. 

In the first frame of an image sequence, $'n'$ tracks are created for $'n'$ detected objects. Then tracks are initialized as follows: each object is assigned a numeric value as "id", second and third field of the tracks are assigned the respective feature vectors of the tracks' corresponding objects, $\widehat{vc}$'s are set to $0$ and Kalman filter fields of the tracks are initialized by centroids of the respective objects. That is,the state vector of the $n^{th}$ track in first frame is initialized as: $s_n=(\widehat{r_n},\widehat{c_n},0,0)$. Along with state vector, The state estimation error covariance matrix ($P$) of the Kalman filter is also initialized as: $P=I_{4\times 4}\ast 100$. From second frame onwards, centroid of each existing track is predicted using the rule (\ref{eqn_21}):
\begin{equation} 
\label{eqn_21}
s_{t|t-1}=As_{t-1|t-1}
\end{equation}
where, $s_{t-1|t-1}$ is the actual state vector of an existing track at the $t^{th}$ time instance and $A=[1000;0100;1010;0101]$ is the state transition model. The state estimation error covariance matrix ($P$) is also updated a priori for each existing track: 
\begin{equation} 
\label{eqn_22}
P_{t|t-1}=AP_{t-1|t-1}A^T+Q
\end{equation}
where, $Q=I_{4\times 4}\ast 0.01$ is the system noise. Then cost of assigning an object detected in the current frame ($v_t$) to an existing track is calculated:
\begin{equation} 
\label{eqn_23}
{cost}_{kl}=
 \begin{cases}
           \frac{1}{3}\sum\limits_{n=1}^{3} |\widehat{g^{k}_{n}}-\widehat{g^{l}_{n}}|, & \text{for $dist\leq (\alpha \widehat{h}\wedge\alpha \widehat{w})$}\\
					\Phi, & \text{Otherwise}
 \end{cases}	
\end{equation}
where, $g^k$'s and $g^l$'s are the intensity distributions of $k^{th}$ track and $l^{th}$ object respectively, ($dist=\sqrt{(\widehat{r_k}-\widehat{r_l})^2+(\widehat{c_k}-\widehat{c_l})^2}$) is the distance between centroids of the same track-object pair, $[\widehat{h},\widehat{w}]$ are the dimensions of the $k^{th}$ track each of which are multiplied by a constant $\alpha=1.5$ and $\Phi$ is a constant which is assigned an arbitrary large value. Combination of cost vectors of length $L$ for all the $K$ number of tracks form the $K\times L$ cost matrix. An object is considered to be associated with a track for which the track has minimum cost. With reference to table \ref{table_I}, ${Object}_1$ has minimum cost for ${Track}_1$; so it is associated with ${Track}_1$. that is, dimensions and intensity distribution of ${Object}_1$ is assigned to the corresponding fields of the ${Track}_1$. Kalman filter of ${Track}_1$ is updated by rule (\ref{eqn_24}) using the centroid ($C_t=[\widehat{r_1},\widehat{c_1}]$) of ${Object}_1$ and the predicted state vector of ${Track}_1$ calculated by rule (\ref{eqn_21}):
\begin{equation} 
\label{eqn_24}
s_{t|t}=s_{t|t-1}+G(C_t-Hs_{t|t-1})
\end{equation}
where, $H=[10;01;00;00]$ is the measurement model which relates state variable to measurement or output variables and $G$ is the Kalman gain calculated as:
\begin{equation} 
\label{eqn_25}
G_t=P_{t|t-1}H^T/(HP_{t|t-1}H^T+R)
\end{equation}
where, $R=I_{2\times 2}$ or unit matrix. This Klaman gain calculated using predicted value of state estimation error covariance matrix ($P$) calculated in (\ref{eqn_22}) is now used to minimize a posterior error covariance matrix:
\begin{equation} 
\label{eqn_29}
P_{t|t}=(I-G_tH)P_{t|t-1}
\end{equation}

\begin{table}[t]
\renewcommand{\arraystretch}{1.3}
\caption{A Sample Cost Matrix For a Video Frame}
\label{table_I}
\centering
\begin{tabular}{c c c c}
\hline
$\textbf{Object}/\textbf{Track}$ & $\textbf{Track}_1$ & $\textbf{Track}_2$ & $\textbf{Track}_3$\\
\hline\hline
 {$\textbf{Object}_1$} & $id = 1$ & $id = 2$ & $id = -1$\\
 & $cost = 1.882$ & $cost = 19.43$ & $cost = \Phi$\\
\hline
$\textbf{Object}_2$ & $id = 1$ & $id = 2$ & $id = -1$\\
 & $cost = 28.79$ & $cost = 4.556$ & $cost = \Phi$\\
\hline
$\textbf{Object}_3$ & $id = -1$ & $id = -1$ & $id = -1$ \\
 & $cost = \Phi$ & $cost = \Phi$  & $cost = \Phi$\\
\hline
\end{tabular}
\end{table}

All the detected object in current frame are associated with an existing track by following the above pursuit. But, some detected object may not be assigned to any of the existing tracks, then these objects are considered as new objects. Such as in table \ref{table_I}, ${Object}_3$ is a new object as it has the arbitrarily high cost ($\Phi$) for all the existing tracks. So, it is assigned a new track. A new track in any intermediate frame is initialized as per the process described previously for the first frame of the image sequence. There may arise another situation where some objects detected in earlier frames (existing track in current frame) may not be associated with any detected object (as ${Track_3}$ in table \ref{table_I}). These undetected objects are either out of the frame or are occluded. In this scenario Kalman filter provides advantage of its predictions. That is centroid of the track is updated by the prior state estimate of the track calculated in (\ref{eqn_21}) and all other fields remain unchanged. At this point, we initialize the $\widehat{vc}$ field of the track to count the number of consecutive frames in which the corresponding object is absent from the view. If this count exceeds a threshold $\eta^{'}$, then we consider the object is permanently out of frame and the corresponding track is removed from the tracker. $\eta^{'}$ is decided as: $\eta^{'}=2\ast \eta$ where, $\eta=$ number of history frames of the current frame. On the other hand, if the occluded object is uncovered in a short while i.e. before its invisibility count  exceeds $\eta^{'}$, then it is again considered for association with the existing track. Throughout our experiments, we have observed that, Klaman prediction of the state of the occluded object or the unassigned track is so accurate that the track is associated with uncovered object correctly. Thus we are able to continue tracking in presence of considerable occlusion.

\section{Algorithm and Performance}
\floatstyle{boxed} % Box...
\restylefloat{figure}
\begin{figure}
\caption{Algorithm for object detection and tracking in complex and variable background}
\label{fig_mainAlg}
\begin{algorithmic}
\State \textbf{Algorithm:} Object detection and tracking in complex and variable background
\\ \State \textbf{Input:} $V$ = a video of size $M\times N\times T$;
 \State \textbf{Output:} $V_{annotated}$ = a video of same size of V marked with Object labels and bounding boxes;
\\ \State Let, $H$ is a set of $\eta$ number of preceding frames of a frame;
\State Let, $Q$ is a set of quantized form of each of $H$;
\begin{enumerate}
\For{$t=\eta+1$ to $T$}
\ \ \ \State Let, $v_t$ is the current frame;
\ \ \ \For{$index=1$ to $\eta$}
\ \ \ \ \ \State Calculate $\overline{v_{t-index}}$ from $v_{t-index}$ using (\ref{eqn_2})-(\ref{eqn_7}) $\Rightarrow H_{index}=\overline{v_{t-index}}$;
\ \ \ \ \ \State Calculate $Q_{index}$ from $H_{index}$ using (\ref{eqn_8}) ; 
\ \ \ \EndFor
\ \ \ \State Intersect each pair of $H_{index}$'s using (\ref{eqn_9}) and unite them using (\ref{eqn_10});
\ \ \ \State Generate history of dissimilarity using (\ref{eqn_11}) and (\ref{eqn_12}) and weight matrix using (\ref{eqn_13});
\ \ \ \State Generate foreground using (\ref{eqn_14}) and detect moving objects in foreground using (\ref{eqn_15});
\ \ \ \State Refine object regions using (\ref{eqn_16})-(\ref{eqn_20});
\If {$f==$ first operating frame of $V$ then}
\ \ \State Initiate each of the ${Track}_k$'s for each of the detected ${Object}_l$'s;
\Else
\ \ \State Calculate cost matrix using (\ref{eqn_23});
\ \ \State Call $Algorithm 1$ (see Appendix 1) using resultant cost matrix as argument;
\ \ \State Annotate the frame $v_t$ using the results returned by $Algorithm 1$;
\EndIf
\EndFor
\end{enumerate}
\end{algorithmic}
\end{figure}

The algorithm detects and tracks moving objects in variable background in three main steps: 1) by estimating and adjusting pseudo motion, 2) by analyzing recent history of a frame and 3) by estimating state vector of an object and solving association problem. The algorithm is depicted in the fig. \ref{fig_mainAlg}.

We have estimated pseudo motion, in form of spatial shift of each of the history frames with respect to the current frame, by estimating phase correlation between them. For this calculation '$\eta+1$' number of Fast Fourier Transforms (FFT) and '$\eta$' number of Inverse Fast Fourier Transforms (IFFT) are done, where '$\eta=4$'. To generate pseudo motion compensated frame '$\eta$' number of IFFT's are done. Each of the FFT's or IFFT's of a frame does $O(MN\log {MN})$ operations, where $'M'$ and $'N'$ are the dimensions of the frame. So, pseudo motion estimation and compensation takes total $O(MN\log {(MN)})$ time. 

At the phase of analysis of recent history of a frame, each of the $\eta$ number of preceding frames have been quantized which takes $O(MN)$ operations, where $'M'$ and $'N'$ are the dimensions of a frame and $\eta=4$. After that we have generated $\eta-1$ number of intersected frames, one weight matrix and one history of dissimilarities matrix by searching the each of quantized frames for values less than '$1$'. The worst case complexity to generate each of those matrices is $O(MN)$. The same number of operations are done while generating the acting background and the foreground. Detecting moving objects in foreground has maximum complexity of $O(MN)$, but as the size of foreground is quite less than the size of frame; number of required comparisons is reduced considerably. Morphological operations on each of the blobs of prospective moving objects take $O(\alpha(\widehat h \widehat w))\approx O(\widehat h \widehat w)$ as $\alpha$ is a constant, where $\widehat h$ and $\widehat w$ are height and width of a blob respectively. If number of detected blobs of prospective moving objects in a frame is $\beta$, then total $O(\beta(\widehat h \widehat w))$ number of operations are required in object region refinement step. So, the total computational complexity of object detection phase is $O(MN)+O(\beta(\widehat h \widehat w))$.

To generate the trajectory of an object we have predicted and updated object state in each frame by Kalman filter and associated each new and previous detection using a cost matrix. The kalman filter equations involve Matrix-vector multiplications and  matrix inversion with $O(n^3)$ operations where $n$ = dimension of the matrix. In our present work, dimension of state vector and other matrices are constants as stated in the previous section. So, the computational complexity of object state prediction and update reduce greatly, say $O(n^3)\rightarrow 1$. To create a cost matrix for $x$ number of existing tracks and $y$ number of detected objects in a frame $O(xy)$. But in average case, most of the track-object cost is not calculated if the centroid distance cost does not meet the threshold criteria. So, in average case the time of cost evaluation reduces drastically.

\section{Experimental Results}
We have conducted our experiments on a computer with an Intel Core i7 3.40 GHz CPU and 16 GB RAM. The algorithm is implemented in Matlab. We have tested the performance
of our algorithm on the benchmark dataset used in \cite{wu2015object}, \cite{kristan2015visual}. This dataset provides ground-truth markings and annotation with various attributes like occlusion, background clutter, rotation, illumination variation etc. on video sequences. We have selected forty video sequences captured by moving camera or have prominent variation in background with other challenges. Each video contains one or two moving object/s in a moderate or very complex variable background. Depending on the major challenges present in environment and background of the videos we have sorted them in following four categories: 1) Occlusion, 2) Rotation (both in-plane and out-of-plane), 3) Deformation and 4) Background clutter. The significant property of our algorithm is that, we did not provide the initial state of object/s in the starting frame; object/s is automatically detected without initialization and training on sample data. The results of moving object detection and tracking are shown in the following part of this section.

To analyze history of commonality and dissimilarity of a frame (say, current frame), we have analyzed $\eta$ number of preceding frames of the current frame where $\eta<<T$ = length of the image sequence. We have applied a heuristic method on our test set of input videos to select the optimum value for $\eta$. This design heuristic is achieved from domain specific information for a large class of problems. We have observed that, if $\eta$ is too small then, may be one or two preceding frames are available to analyze the historical information of the current frame. Now, if the image sequence contains articulated object which has partial movement in these preceding frames, then no significant information on position and size of object region can be extracted from these history frames. On the other hand, if object has full-body movement in each consecutive frames, then it displays vast amount of displacement in the series of preceding frames. Thus, if $\eta$ is too large (say, six or ten), then also estimation of position and size of object region becomes imprecise. So in our present work, we have concluded $\eta=4$ as an optimum size of history of the current frame.

We have set the value of another parameter $\alpha$ from experience to define the search region around a moving object. We can say intuitively that, objects with uniform velocity maintain uniform amount of shift of object centroid in subsequent frames. So, searching an object of current frame with uniform motion within a region $(\widehat h* \widehat w)$ in the next frame is sufficient. But, if the object has abrupt motion then an increased search region like, $\alpha(\widehat h*\widehat w)$ is required. We have observed through our experiment that, increment of search region by less than twice of its original size (i.e. $\alpha<2$) is sufficient to cover up the abrupt displacement of an object. So, $\alpha=1.5$ is set as optimum value to define the size of search region around an object.

\begin{table*}[t]
\renewcommand{\arraystretch}{1.3}
\caption{Summary of Experimental Result on Selected Image Sequences}
\label{table_II}
\centering
\begin{tabular}{c c c c c}
\hline
\textbf{Sequence Category} & \textbf{Mean FPS} & \textbf{Mean TD} & \textbf{Mean FD} & \textbf{Mean MD}\\ 
\hline\hline
Occlusion (23 sequences) &	275	& 83.0\% &	3.12\% &	0\% \\ 
Rotation (both in-plane and out-of-plane - 27 sequences) & 294	& 90.97\% &	2.17\%	& 0\%\\
Deformation (25 Sequences) &	211	& 74.1\% &	8.94\%	& 3.57\%\\
Background clutter (22 sequences) &	153 &	64.2\% &	10.69\%	& 4.74\%\\ 
\hline
\end{tabular}
\end{table*}

We have quantitatively evaluated our algorithm using four parameters: Frames per second (FPS), True Detection (TD), False Detection (FD) and Missed Detection (MD). FPS is the count of annotated frames displayed per second. TD is calculated as the percentage of frames with successful object detection and tracking in each sequence : 
\begin{equation}
\label{eqn_26}
TD=\frac{n_{td}}{N}\times100
\end{equation}
 where N = Total number of frames in an image sequence and $n_{td}$ = number of frames with truly detected object. We have measured the success of a frame as per the following rule- if in a frame, bounding box around a detected and tracked object overlaps with the bounding box of the ground truth i.e. $|C_t-C_g|\leq B_t/B_g$, where C's are the centroids and B's are the bounding boxes; then the frame is marked as a successful frame. If object is detected in a position of a frame where ground truth does not indicate any or detected bounding box does not overlap with the bounding box of the ground truth i.e. $|C_t-C_g|>B_t/B_g$; then the detection is considered as false detection and formulated as: 
\begin{equation}
\label{eqn_27}
FD=\frac{n_{fd}}{n_{td}+n_{fd}}\times100
\end{equation} 
If object is not detected in a frame, but ground truth value for that frame exists; then the situation is considered as Missed Detection and formulated as: 
\begin{equation}
\label{eqn_28}
MD=\frac{n_{md}}{n_{td}+n_{md}}\times100
\end{equation} 
We have grouped the selected image sequences into four categories as mentioned above and  calculated all the four metrics for individual sequence of each category. Then the mean value of each metrics is computed for each category. The summary of experimental result in Table \ref{table_II} shows that the system is highly successful in detecting moving object in a variable background and tracking the object throughout the input sequence for most of the test sequences. However, the system is not that efficient when the input sequence contains many small objects against very complex background. The primary reason of such poor performance is the failure of the object detection part and sometimes due to the failure of the tracking part.

\begin{figure*}[t]
\centering
\subfloat[]{\includegraphics[width=1.5in]{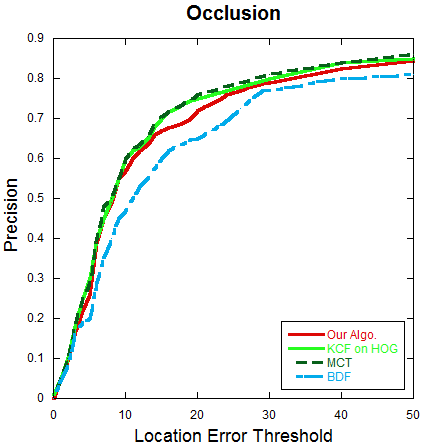}
\label{fig_go}}
\hfil
\subfloat[]{\includegraphics[width=1.5in]{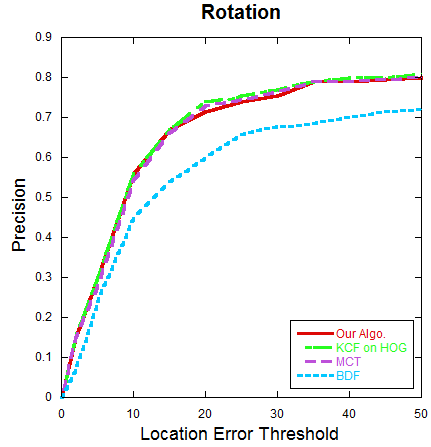}
\label{fig_gr}}
\hfil
\subfloat[]{\includegraphics[width=1.5in]{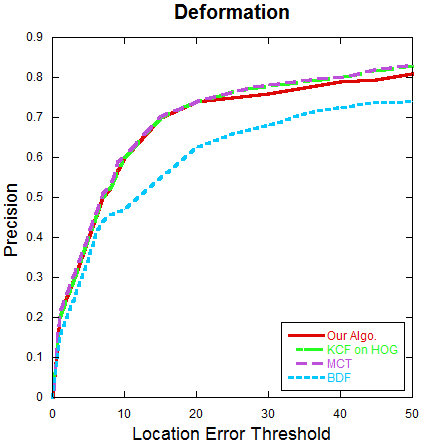}
\label{fig_gd}}
\hfil
\subfloat[]{\includegraphics[width=1.5in]{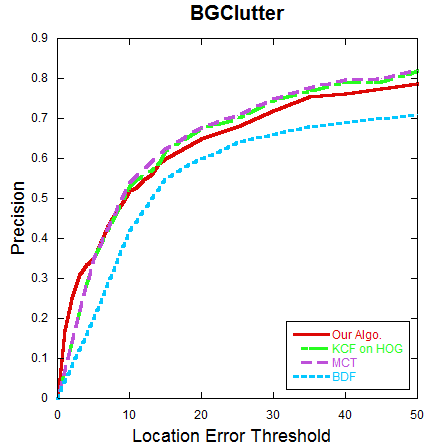}
\label{fig_gb}}
\caption{Precision curves for all the selected test sequences with attributes: Occlusion, Rotation (both in-plane and out-of-plane), Deformation and Background clutter.}
\label{fig_5}
\end{figure*}

\begin{figure}[t]
\centering
\includegraphics[width=3in]{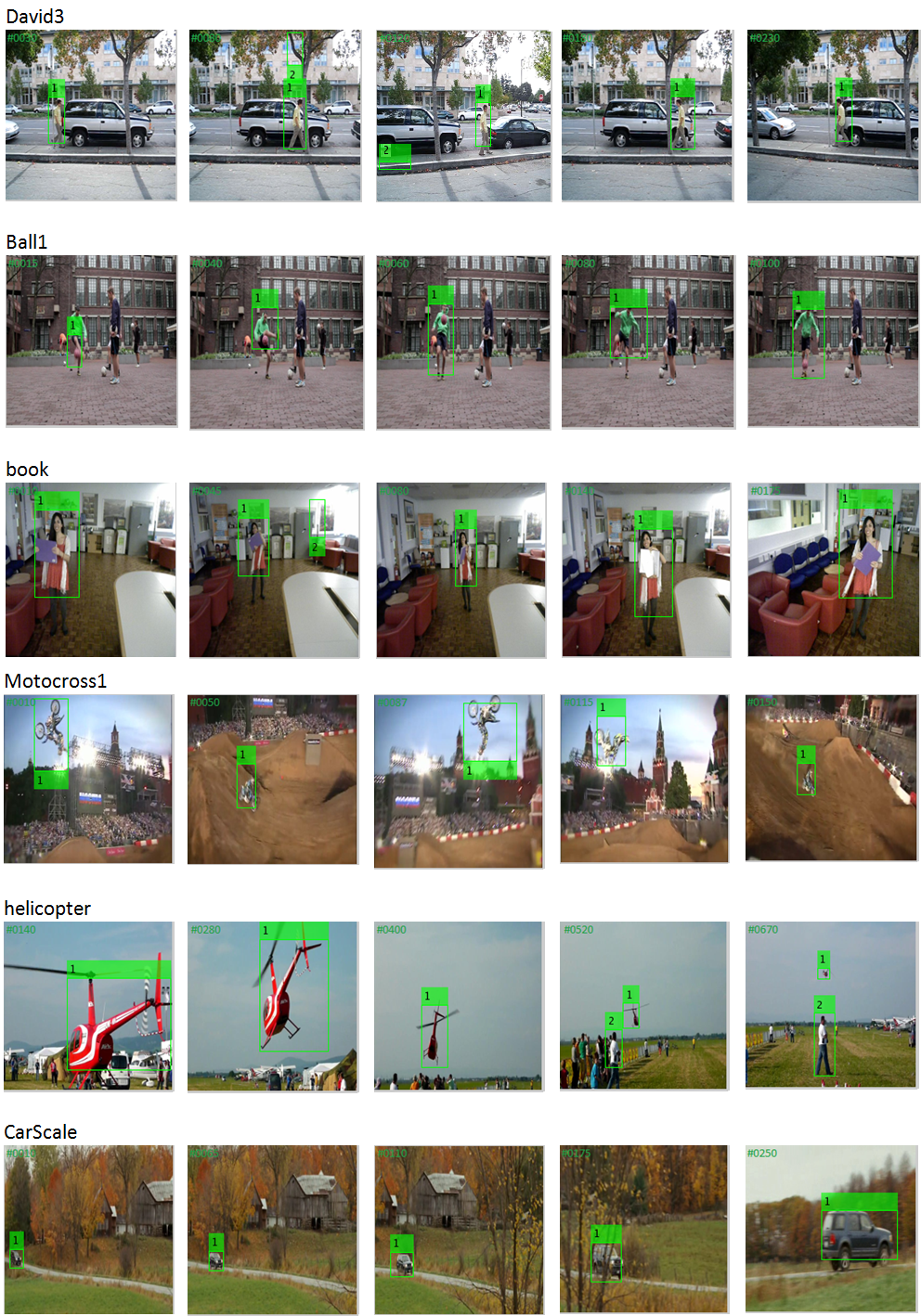}
\caption{Results of Object detection and tracking by proposed algorithm.}
\label{fig_6A}
\end{figure}

\begin{figure}[t]
\centering
\includegraphics[width=3in]{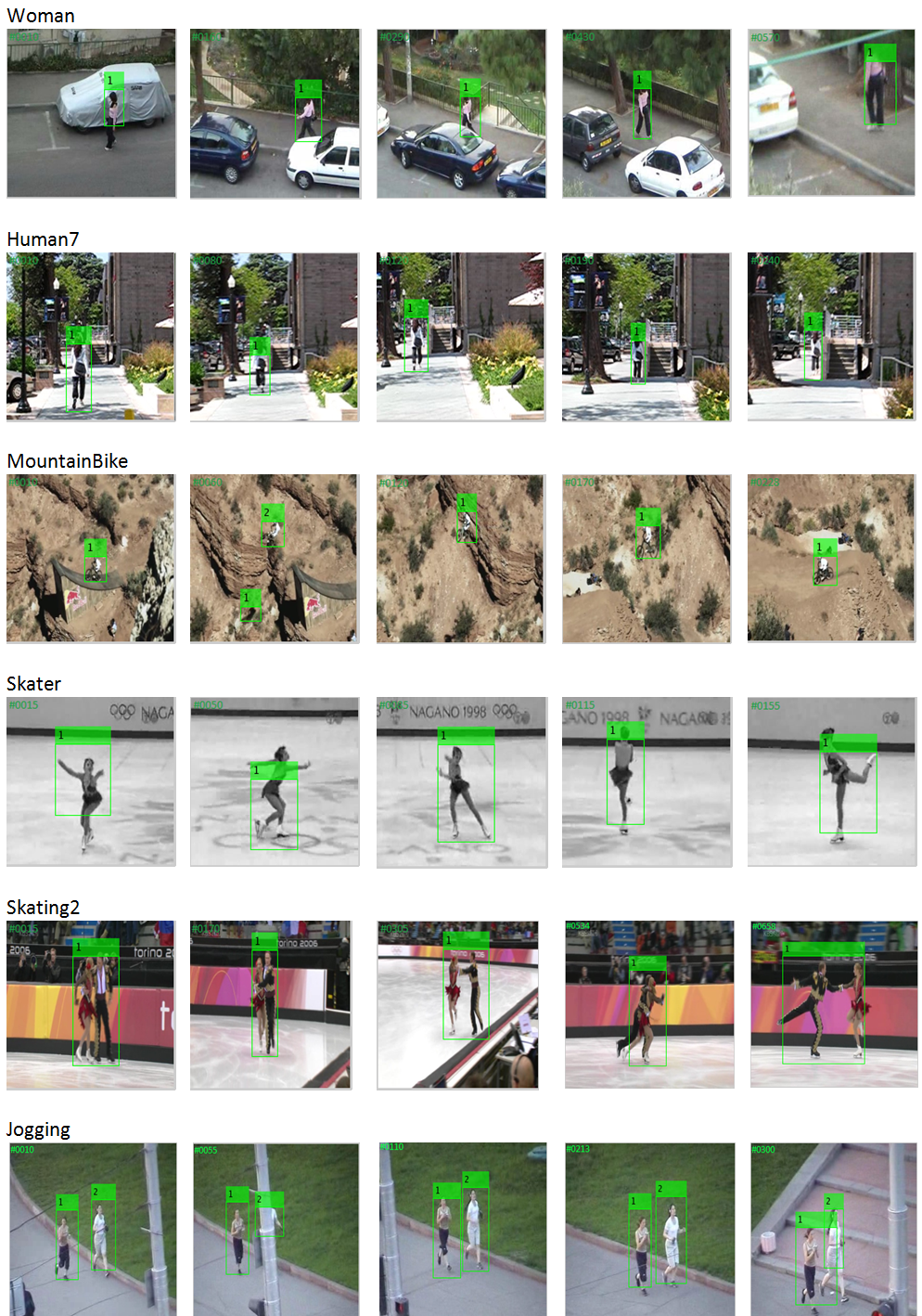}
\caption{Some more results of Object detection and tracking by proposed algorithm.}
\label{fig_6B}
\end{figure}

Fig. \ref{fig_6A} and Fig. \ref{fig_6B} contains the results of object detection and tracking in widely spaced sequential frames of some of the input image sequences. Each row of the figure contains result for different image sequences. Our proposed algorithm is very robust in detecting object against the challenge of rotation (both in-plane and out-of-plane). It also performs well in handling occlusion as depicted in Woman, David3 and Jogging sequences. Our proposed method performs satisfactorily in presence of deformation of non-rigid object as illustrated in sequences like Skater, skating2, jogging etc. It also detects and tracks object/s very well in cluttered background with gradual illumination variation as presented in MountainBike, David3, CarScale, Human7 etc. When a group of people are moving in a close proximity with similar speed; the proposed method detects and tracks them as one single object as shown in the sequence Skating2. We have compared our algorithm with three state-of-the-art algorithms: Tracking using discriminative motions models based on Particle filter framework- motion context tracker (MCT) \cite{duffner2016using}, High-Speed Tracking with Kernelized Correlation Filters using Histogram of Oriented Gradients (KCF on HOG) \cite{henriques2015high}, a short-term tracking algorithm -Best Displacement Flow (BDF) \cite{maresca2014clustering}. We have used precision curve \cite{wu2015object}, \cite{henriques2015high} as the evaluation metric for the comparison. The precision curve express the percentage of frames in which the detected object centroid is within a given threshold of the ground truth position. We presented the comparison on performance of all the four algorithms on selected input sequences with variable background; categorized in four groups  in Fig. \ref{fig_5}: 1) Occlusion, 2) Rotation (both in-plane and out-of-plane), 3) Deformation and 4) Background clutter. The proposed algorithm exhibits comparable results with KCF on HOG and MCT in handling all the four selected nuisance. Our proposed method also have shown superiority against BDF in handling all of the four situations except occlusion.

\floatstyle{plain}
\restylefloat{figure}

\section{Conclusion and Future Work}
The algorithm proposed in this paper efficiently detects and tracks one or more moving object/s simultaneously in variable background. The proposed algorithm has several advantages: 1) all the experimented image sequences are captured by one moving camera without any additional sensor, 2) The major advantage of the algorithm is that it does not depend on any prior knowledge of the environment (scene) or any information about the shape of the objects to be detected, 3) The significant achievement of the proposed algorithm is that, it does not require object region initialization at the first frame or training on sample data to perform. We have obtained satisfactory results on benchmark videos containing variable background \cite{wu2015object}, \cite{kristan2015visual}. Performance of our algorithm is also comparable and sometimes superior with respect to the state-of-the-art methods \cite{duffner2016using}, \cite{henriques2015high}, \cite{maresca2014clustering}.  

The algorithm has compensated the pseudo motion in background due to camera motion very efficiently by employing phase correlation method. The proposed method of calculating commonality in recent history of current frame, also models the acting background very efficiently. By experiments so far, we have observed that, frame differencing between background model and the current frame generates the foreground with minimum noise and false detection. The idea of using history of dissimilarities of the current frame also worked efficiently to detect actual moving object accurately from the foreground. The morphological operation proposed in the algorithm, produced the optimum object region which enabled the tracker to track very closely situated objects separately. The proposed algorithm has successfully implemented the Kalman filter to track moving objects and to handle occlusion. The experimental results show that, use of two features- dimensions of the object and its intensity distribution has very efficiently solved the problem of data association during tracking. In future, we would like to extend our work to detect and track object in a very crowded scene or in presence of extreme illumination variation.

% or


\appendices
\section{}
Algorithm to update feature vector of each track is $Algorithm 1$, which is depicted in Fig. \ref{fig_alg1}.
\floatstyle{boxed} % Box...
\restylefloat{figure}
\begin{figure}
\caption{Algorithm to update feature and state vector of each track }
\label{fig_alg1}
\begin{algorithmic}
\State \textbf{Algorithm 1:} Updating feature and state vector of tracks  
\\ \State \textbf{Input: } 
\State ${cost}_{K\times L}$- cost matrix for the frame $v_t$
\State \textbf{Output: } Updated feature and state vector of track for the frame $v_t$
%\\ \State 
\begin{enumerate} 
\For{$k=$ each $Track_k$}
\ \ \If{$cost_{kl}==min(cost_{k1}...cost_{kL})$}
\ \ \ \ \State Update $Track_k$ by the features of $Object_l$ and state vector calculated using (\ref{eqn_24})-(\ref{eqn_29});
\ \ \ \ \State Remove entry of $Object_l$ from cost matrix;
\ \ \EndIf
\EndFor
\ \ \State Label track with null entry in cost matrix as unassigned;
\ \ \State Label remaining object in cost matrix as candidateNew;
\For{$i=$ each unassigned track}
\ \ \ \State $Track_i$ is updated by state prediction calculated using (\ref{eqn_21}), (\ref{eqn_22});
\EndFor
\For{$j=$ each candidateNew}
\ \ \If{$min(cost_{k1}...cost_{kL})==\Phi$} 
\ \ \ \ \State assign $Object_j$ to a new track;
\ \ \ElsIf{$min(cost_{k1}...cost_{kL})<\Phi$} 
\ \ \ \ \State $Object_j$ is discarded as erroneous detection;
\ \ \EndIf
\EndFor 
\end{enumerate}
\State $Return$ updated feature and state vector for each track;
\end{algorithmic}
\end{figure}

%\section{}

%\section*{Acknowledgment}


\ifCLASSOPTIONcaptionsoff
  \newpage
\fi



\bibliographystyle{IEEEtran}


\begin{IEEEbiography}[{\includegraphics[width=1in,height=1.25in,clip,keepaspectratio]{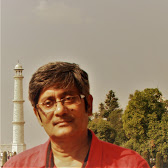}}]{Kumar S. Ray}
, PhD, is a Professor in the Electronics and Communication Sciences Unit at Indian Statistical Institute, Kolkata, India. He is an alumnus of University of Bradford, UK. Prof. Ray was a member of task force committee of the Government of India, Department of Electronics (DoE/MIT), for the application of AI in power plants. He is the founder member of Indian Society for Fuzzy Mathematics and Information Processing (ISFUMIP) and member of Indian Unit for Pattern Recognition and Artificial Intelligence (IUPRAI). 
His current research interests include artificial intelligence, computer vision, commonsense reasoning, soft computing, non-monotonic deductive database systems, and DNA computing. He is the author of two research monographs viz, Soft Computing Approach to Pattern Classification and Object Recognition, a unified conept, Springer, Newyork, and Polygonal Approximation and Scale-Space Analysis of closed digital curves, Apple Academic Press, Canada, 2013.
\end{IEEEbiography}

\begin{IEEEbiography}[{\includegraphics[width=1in,height=1.25in,clip,keepaspectratio]{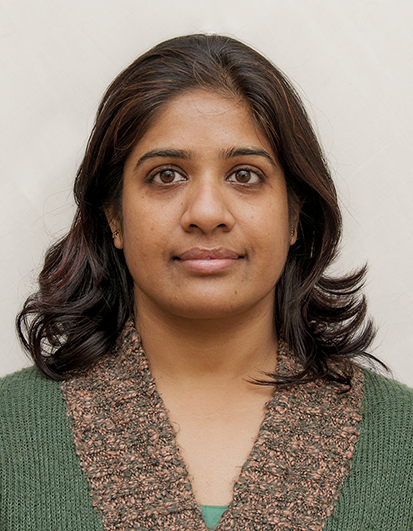}}]{Soma Chakraborty}
is working as a project linked scientist at Electronics and communication Sciences Unit of Indian Statistical Institute, Kolkata, India. Prior to joining ISI in December 2013, she worked in software industry as developer in healthcare and manufacturing domain for 7.8 years. She received her bachelor’s degree in Information Technology from West Bengal University of Technology, India in 2005 and MS degree in Software System from BITS, Pilani, India in 2012. Her research interest includes video image processing, computer vision and pattern recognition.
\end{IEEEbiography}

%\newpage



%\vfill


\end{document}